# Automated Seam Folding and Sewing Machine on Pleated Pants for Apparel Manufacturing

Ray Wai Man Kong[1]

[1] Adjunct Professor, City University of Hong Kong, Hong Kong

[1] Modernization Director, Eagle Nice (International) Holding Ltd, Hong Kong

DOI: https://doi.org/10.5281/zenodo.16607787

Published Date: 30-July-2025

*Abstract:* The applied research is the design and development of an automated folding and sewing machine for pleated pants. It represents a significant advancement in addressing the challenges associated with manual sewing processes. Traditional methods for creating pleats are labour-intensive, prone to inconsistencies, and require high levels of skill, making automation a critical need in the apparel industry. This research explores the technical feasibility and operational benefits of integrating advanced technologies into garment production, focusing on the creation of an automated machine capable of precise folding and sewing operations and eliminating the marking operation.

The proposed machine incorporates key features such as a precision folding mechanism integrated into the automated sewing unit with real-time monitoring capabilities. The results demonstrate remarkable improvements: the standard labour time has been reduced by 93%, dropping from 117 seconds per piece to just 8 seconds with the automated system. Similarly, machinery time improved by 73%, and the total output rate increased by 72%. These enhancements translate into a cycle time reduction from 117 seconds per piece to an impressive 33 seconds, enabling manufacturers to meet customer demand more swiftly. By eliminating manual marking processes, the machine not only reduces labour costs but also minimizes waste through consistent pleat formation. This automation aligns with industry trends toward sustainability and efficiency, potentially reducing environmental impact by decreasing material waste and energy consumption.

Moreover, the system's scalability allows operators to manage multiple machines, further boosting productivity. This adaptability makes it a versatile solution for various manufacturers, accommodating different production needs effectively. The research underscores the transformative potential of automated seam folding and sewing machines in the apparel industry, providing valuable insights for manufacturers seeking to adopt cutting-edge technologies. By streamlining operations and improving consistency, this system positions manufacturers at the forefront of innovation, enabling them to meet evolving consumer demands and remain competitive in a rapidly changing market. The concrete data on efficiency improvements underscores the machine's value, offering clear benefits for manufacturers seeking to adopt cutting-edge technologies and enhance their competitive edge.

## I.  INTRODUCTION

In the rapidly evolving landscape of the apparel industry, the integration of automation technologies has become essential for enhancing productivity, improving quality, and reducing operational costs. One of the most significant advancements in this domain is the development of automated seam folding and sewing machines. These innovative machines streamline the garment manufacturing process by automating critical tasks that were traditionally performed manually, thereby increasing efficiency and consistency in production.

The benefits of automated seam folding and sewing machines are manifold. Firstly, they significantly reduce labour costs by minimizing the need for manual intervention in repetitive tasks, allowing manufacturers to allocate human resources to more complex and value-added activities. Secondly, these machines enhance precision and accuracy in seam folding and sewing, resulting in higher-quality finished products with fewer defects. This quality improvement not only boosts customer satisfaction but also reduces waste associated with rework and returns.

Moreover, automated seam folding and sewing machines contribute to faster production cycles. By accelerating the folding and sewing processes, manufacturers can respond more swiftly to market demands and changing consumer preferences, ultimately leading to shorter lead times and improved competitiveness. Additionally, the integration of advanced technologies such as AI and machine learning into these machines allows for real-time monitoring and optimization of production processes, further enhancing operational efficiency.

The decision to conduct applied research on automated seam folding and sewing machines stems from the clothing manufacturing need for innovation in the apparel manufacturing sector. As global competition intensifies and consumer expectations evolve, manufacturers must adopt cutting-edge technologies to remain relevant and profitable. Our applied research aims to explore the technical feasibility, operational benefits, and potential challenges associated with implementing these automated systems in garment production. By investigating the impact of automated seam folding and sewing machines on overall manufacturing efficiency, we seek to provide valuable insights that can guide industry stakeholders in their pursuit of automation and continuous improvement.

In short, the advent of automated seam folding and sewing machines represents a pivotal advancement in apparel automation, offering substantial benefits in terms of cost reduction, quality enhancement, and production efficiency. Our applied research endeavours to illuminate the transformative potential of these technologies, paving the way for a more innovative and competitive apparel industry.

## II. PLEATED PANTS STYLES

Pleated pants have long been a staple in men's fashion, offering a blend of sophistication and comfort that makes them suitable for various occasions. The slightly tailored cut of pleated pants provides a sharp silhouette while ensuring freedom of movement, making them an ideal choice for both casual and formal settings. The design features, such as the extended button tab, add character and elevate even the simplest looks, allowing wearers to express their personal style effortlessly.

For instance, the Toso Beige Pleated Pants exemplify this balance of style and functionality. Their tailored fit enhances the overall aesthetic while maintaining comfort, making them versatile enough for both work and leisure. The pleats not only contribute to the visual appeal but also provide additional room, ensuring ease of movement throughout the day.

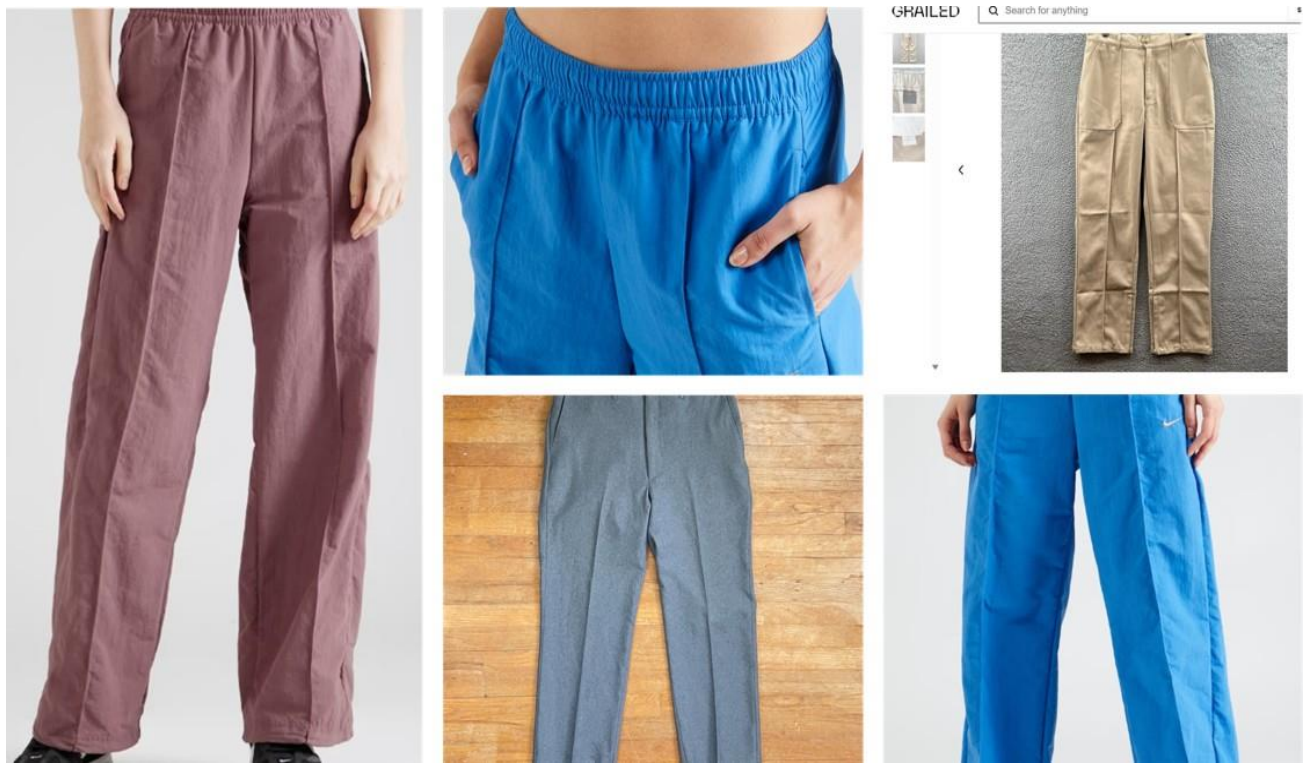

**Figure 1: Pleated Pants Diagram from Nike, Grailed and Etsy**

Similarly, the Nike Men's Apparel Tech Fleece Tailored Pants showcase in Fig. 1 how pleated designs can be adapted to modern, sporty aesthetics. Crafted from a premium French Terry cotton blend, these pants combine a loose, streamlined fit with practical features like a drawcord waist and front seam detail. The tonal logo embroidery adds a touch of branding without overwhelming the overall design, making it perfect for upgrading casual wear.

Pleated pants represent a harmonious fusion of style, comfort, and practicality. Whether in a classic tailored version or a contemporary athletic style, they remain a versatile choice for those looking to enhance their wardrobe with pieces that offer both elegance and ease.

The pleated trouser for women looks nice with a nice blouse and heels or flats. It's a polished look, and it's classic style for women's dress pants. The design has been added a bit of fullness through the upper hip and replaced the darts with pleats. This doesn't make the pants loose or oversized; it just makes the fit a bit more relaxed through the upper hip.

The leg shape in the design remains the same as the block. Alter the leg shape as well (adding a pleat to a tapered or flared leg). The specific styles have been provided; add the pleat before changing the leg shape. The manufacturing of pants to make the front long pleat is not easy to build it straight without bends or curvature in Fig. 2.

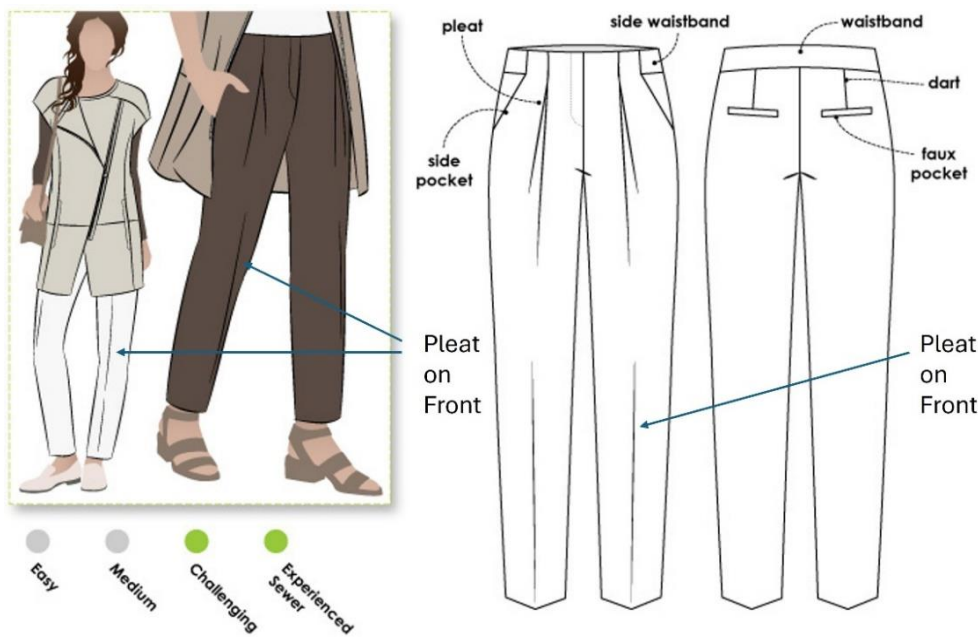

**Figure 2: Pleated Pants Pattern Diagram, source from the clothingpatterns101**

The clothing process for a long pleat in front of pants requires folding the front fabric piece and sewing a single sewing line. The folding dimension is so tight with narrow tolerance. It is so difficult to do it in manual operation in Fig. 3.

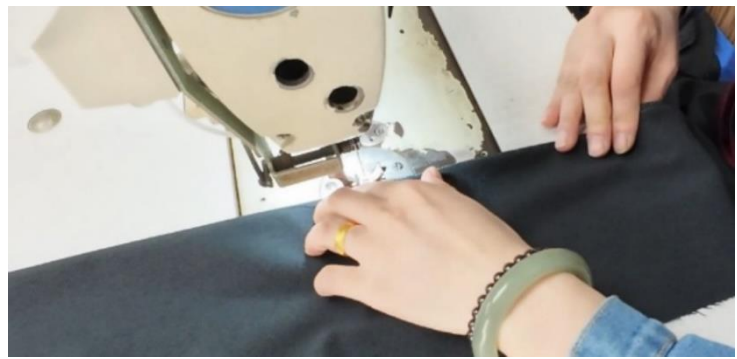

**Figure 3: Operator manual folding and sewing operation**

Therefore, the sewing process typically needs to be driven the automation. In the market, there is no specific machine to fold and sew the pleat on fabric at the front of pants. The operation is manual work. The operator needs to draw a line on fabric and then manually fold and sew with the single needle sewing machine.

#### A. The Challenges of Manual Sewing Operations for Long Pleats in Pants

The clothing process for creating long pleats in the front of pants is a meticulous task that requires precision and skill. This process involves folding the front fabric piece and sewing a single line to secure the pleat in place. However, the folding dimensions are often tight, with narrow tolerances that demand a high level of accuracy. Achieving this level of precision through manual operation can be exceedingly challenging, as illustrated in Figure 3, which depicts an operator engaged in the manual folding and sewing operation.

<u>**The Complexity of Manual Pleat Sewing**</u>

Manual sewing operations for long pleats present several difficulties. First and foremost, the operator must ensure that the fabric is folded evenly and consistently along the designated line. This requires not only a steady hand but also a keen eye for detail. Any deviation from the specified dimensions can result in uneven pleats, which can compromise the overall aesthetic of the garment. The challenge is further compounded by the need to avoid curves or bends in the fabric piece during the folding process. Even the slightest misalignment can lead to puckering or distortion, which is particularly problematic in high-quality garments where visual appeal is paramount.

Additionally, the manual operation involves drawing a line on the fabric to guide the folding process. This step is crucial, as it serves as a reference point for the operator. However, drawing the line accurately can be time-consuming and requires a level of skill to ensure that it is straight and correctly positioned. Once the line is drawn, the operator must carefully fold the fabric along this line and then secure it with a single needle sewing machine. This process demands significant dexterity and concentration, as the operator must maintain tension on the fabric while feeding it through the machine. Any lapse in focus can lead to mistakes, resulting in wasted fabric and increased production costs.

#### B. The Need for Automation

Given the complexities and challenges associated with manual sewing operations for long pleats, there is a compelling case for automation in this process. Currently, the market lacks specific machines designed to fold and sew pleats on fabric at the front of pants, leaving operators to rely on manual methods. This gap in the market highlights the need for innovative solutions that can streamline the pleating process and enhance efficiency.

Automated sewing machines designed for this purpose could significantly reduce the labor intensity of the operation while improving accuracy and consistency. By integrating advanced technologies such as sensors and programmable controls, these machines could ensure precise folding and sewing, eliminating the variability associated with manual operations. Furthermore, automation would allow for faster production cycles, enabling manufacturers to meet the increasing demand for high-quality garments without compromising on craftsmanship.

The manual sewing operation for creating long pleats in pants is fraught with challenges that can hinder efficiency and quality. The tight folding dimensions and the need to avoid curves and bends in the fabric make this process particularly difficult for operators. As the apparel industry continues to evolve, the introduction of automated solutions for pleat folding and sewing will be essential in addressing these challenges, ultimately leading to improved production outcomes and enhanced garment quality.

### III. LECTURE REVIEW

In the context of modern manufacturing, the implications of innovative technology are profound, particularly concerning the growth of automation and engineering technology. From Prof. Ray Wai Man Kong's articles [1] [2] [3] [4] [5] [6] [7] [8] [9] [10] [11] [12] [13] [14], his team can explore how these advancements are reshaping the manufacturing landscape and driving efficiency, productivity, and competitiveness.

#### A. Automated Machinery Development in Clothing Manufacturing: Embracing Innovation for the Future

The fashion industry has long been a cornerstone of global manufacturing, with clothing production playing a vital role in meeting the ever-evolving demands of consumers. However, as the world progresses toward automation and Industry 4.0, the traditional methods of garment manufacturing are being challenged to adapt or risk becoming obsolete. Automated machinery development is at the forefront of this transformation, offering solutions that enhance efficiency, precision, and scalability while addressing the growing need for sustainable practices.

The above articles from Prof Dr Ray Wai Man Kong explore the advancements in automated machinery within the clothing manufacturing sector, focusing on innovative technologies such as vacuum suction grabbing systems, AI-driven intelligent learning, and line balancing strategies. By examining these developments, he aims to highlight how automation is reshaping the industry and enabling manufacturers to stay competitive in a rapidly changing market.

### B. Vacuum Suction Grabbing Technology: Revolutionizing Garment Handling

One of the most promising innovations in garment automation is vacuum suction grabbing technology. Traditionally, handling fabric pieces has been a labour-intensive process prone to errors due to the delicate nature of textiles. However, Prof Dr Ray WM Kong et al. (2024) introduced a novel vacuum suction grabbing system designed specifically for grasping fabric pieces with precision and minimal damage. This technology utilises advanced sensors and pneumatic systems to ensure accurate positioning and handling of materials, significantly reducing waste and improving production efficiency.

The application of this technology in automated sewing machines has been particularly impactful. For instance, Prof Dr Ray WM Kong et al. (2025) developed an innovative pulling gear mechanism for automated pant bottom hem sewing machines, which integrates vacuum suction grabbing to enhance the accuracy and speed of the sewing process. By automating these tasks, manufacturers can achieve higher productivity while maintaining quality standards.

### C. AI-Driven Intelligent Learning in Manufacturing Automation

Artificial intelligence (AI) is another transformative force in garment manufacturing automation. Prof Dr Ray WM Kong et al. (2025) explored the potential of AI-driven intelligent learning systems to optimize production processes and improve decision-making. These systems leverage machine learning algorithms to analyze vast amounts of data, enabling machines to adapt to new tasks and optimize performance over time.

For example, AI-powered conveyor systems, such as the magnetic levitation (Maglev) conveyors developed by Prof Dr Ray WM Kong (2025), offer a revolutionary approach to material handling in automated assembly lines. These systems use electromagnetic fields to suspend and transport materials, eliminating friction and reducing wear and tear on machinery. This innovation not only enhances efficiency but also contributes to energy savings and longer machine lifespans.

Moreover, AI-driven systems can be integrated with smart sensors to monitor production processes in real time. By analyzing data from these sensors, manufacturers can identify bottlenecks, predict maintenance needs, and optimize resource allocation, ultimately leading to more sustainable and cost-effective operations.

### D. Line Balancing Strategies for Optimal Production Efficiency

In garment manufacturing, achieving optimal line balancing is critical to ensuring smooth production flows and minimizing idle time. Prof Dr Ray WM Kong et al. (2025) developed a mixed-integer linear programming (MILP) model specifically designed to address the complexities of line balancing in modern garment lines. This model considers various factors, such as task durations, machine capacities, and worker skills, to create balanced production lines that maximize efficiency.

### E. Collaboration Between Hong Kong and China: A Synergistic Approach

Hong Kong and China have emerged as key players in the development of innovative technologies for the fashion industry. Hong Kong's strengths in research and development (R&D), coupled with its advanced infrastructure, provide a fertile ground for technological advancements. Meanwhile, China's vast manufacturing capabilities and market reach enable the rapid implementation of these innovations on an industrial scale.

Prof Dr Ray WM Kong et al. (2025) emphasized the importance of collaboration between Hong Kong and China in driving technological progress within the garment industry. By leveraging Hong Kong's expertise in automation technologies and China's production capacity, both regions can create a synergistic ecosystem that fosters innovation and drives global competitiveness.

### Challenges and Future Directions

Despite the promising advancements in automated machinery, several challenges remain. The high initial costs of implementing advanced technologies, such as AI-driven systems and vacuum suction grabbing mechanisms, pose a barrier for smaller manufacturers. Additionally, integrating these technologies with existing production lines requires significant effort and expertise to ensure seamless operation.

Looking ahead, the industry must focus on overcoming these barriers by investing in R&D and fostering partnerships between academia, government, and private enterprises. As highlighted by Prof Dr Ray WM Kong (2025), future innovations should prioritize sustainability, with a focus on reducing waste, energy consumption, and environmental impact.

The development of automated machinery is revolutionizing the clothing manufacturing industry, offering solutions that enhance efficiency, precision, and scalability while addressing the growing need for sustainable practices. Innovations such as vacuum suction grabbing technology, AI-driven intelligent learning systems, and advanced line balancing strategies are paving the way for a more competitive and environmentally friendly future.

By fostering collaboration between regions like Hong Kong and China, manufacturers can leverage their respective strengths to drive technological progress and stay ahead in the global market. As the industry continues to evolve, embracing these advancements will be crucial for meeting the demands of modern consumers and ensuring long-term success in the fashion sector.

Referred to Domskiene, J et al. Sewing, the adhesive bonding and seam sealing technologies are used to obtain e-textile packages with woven and knitted conductive textiles. Produced e-textile packages described in terms of thickness, bending rigidity and general appearance. The exploitation properties of prepared samples were tested by a cycle tensile experiment and discussed on the basis of the variation of linear electrical resistance property.

**F.  Pull Force Analysis in Clothing Manufacturing Machinery Development**

In the context of clothing manufacturing, pull force analysis is a critical aspect of developing reliable and effective automated machinery. This analysis focuses on understanding and optimizing the forces involved when materials are moved through various stages of production. Here's an organized overview of how pull force analysis plays a role in this field:

**Definition and Importance:**

Pull force refers to the amount of effort or energy required to move materials within a machine.

Proper analysis ensures that machinery operates efficiently, handles materials without damage, and maintains durability.

**Key Applications:**

Vacuum Suction Grabbing Technology: This technology is used to handle delicate fabrics. Pull force analysis here involves determining the optimal vacuum pressure needed to grip fabric securely without causing tears or damage.

**Automated Sewing Machines:**

These machines pull fabric through for stitching. Analyzing pull force ensures even stitching and prevents fabric edge damage, enhancing product quality and machine longevity.

**Conveyor Systems:**

Used to transport garments or materials along production lines, these systems require optimal pull force to avoid jams and ensure smooth operation.

**Factors Influencing Pull Force:**

- Weight of the material
- Friction between moving parts
- Speed of movement
- Type of machinery used

**Methods of Analysis:**

Testing different scenarios to observe effects on machinery and materials.

Utilizing simulations to predict outcomes under various pull force conditions.

**Benefits of Proper Pull Force Analysis:**

Enhances efficiency in production processes.

Prevents unnecessary wear and tear on equipment.

Reduces the risk of product damage, ensuring higher quality output.

According to the laws of friction, the frictional force *F* is proportional to the normal force or normal load *FN*. The effect of the normal force is such that friction increases with it. This law can be represented in the form of an equation.

$F \propto F_N$ (1)

Or, $F = \mu F_N$ where unit of frictional force: Newton or N, Dimensions of frictional force: MLT-2

This equation gives the magnitude of the frictional force. Its direction is opposite to the applied force. The constant of proportionality μ is called the friction coefficient or coefficient of friction (COF). Unlike other proportionality constants, it does take a constant value and has no unit. Its value depends upon the two surfaces in contact.

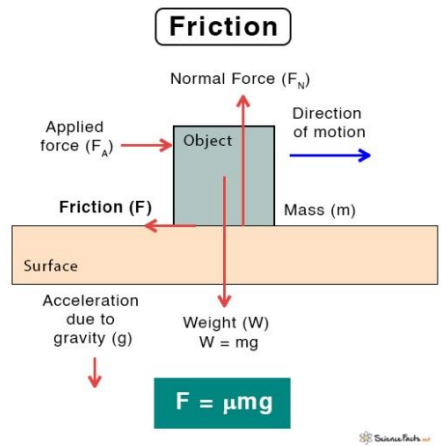

**Figure 4: Friction Force Schematic Diagram**

Pull force analysis is essential to resist the friction force in Fig. 4 (source: sciencefacts.net) between the fabric and the table, which is a designed machinery that operates smoothly, handles materials effectively, and maintains high production standards in clothing manufacturing. By integrating pull force analysis into the development of automated systems, manufacturers can achieve more efficient, durable, and high-quality production processes, ultimately contributing to the industry's success and competitiveness.

## IV. DESIGN OF AUTOMATED FOLDING AND SEWING MACHINE FOR PLEATED PANTS STYLES

### A. Understanding the Challenges

As previously discussed, the manual operation of folding and sewing pleats involves several complexities. Operators must ensure precise alignment and consistent folding along a designated line, which is difficult to achieve without the risk of misalignment or distortion. The manual process is labour-intensive, requiring significant skill and concentration, which can lead to variability in the final product. Additionally, the reliance on manual methods can result in increased production times and costs, making it imperative to explore automation solutions.

### B. Key Features of an Automated Folding and Sewing Machine

Precision Folding Mechanism:

The core of the automated machine must include a precision folding mechanism capable of accurately folding the fabric along predetermined dimensions. This could involve the use of adjustable guides and sensors that ensure the fabric is folded evenly and consistently, minimizing the risk of curves or bends. The mechanism should be designed to accommodate various fabric types and thicknesses, allowing for versatility in production.

Integrated Line Drawing System:

To eliminate the need for manual line drawing, the machine could incorporate an integrated system that uses laser or digital projection technology to mark the folding line directly onto the fabric. This feature would enhance accuracy and save time, ensuring that operators can quickly set up the machine for different styles and designs.

Automated Sewing Unit:

The sewing unit should be equipped with advanced technology to ensure precise stitching along the folded pleat. This could include programmable sewing patterns that allow for adjustments based on fabric type and desired pleat depth. Additionally, the sewing unit should feature automatic tension control to maintain consistent fabric feed and prevent puckering during the sewing process.

Real-Time Monitoring and Feedback:

Incorporating sensors and monitoring systems would enable the machine to provide real-time feedback on the folding and sewing processes. This feature would allow for immediate adjustments to be made if any deviations from the desired specifications are detected, ensuring high-quality output and reducing waste.

User-Friendly Interface:

To facilitate ease of use, the machine should be equipped with an intuitive user interface that allows operators to easily input specifications, select sewing patterns, and monitor production progress. Training operators on the new technology should also be a priority, ensuring they can effectively utilize the machine's capabilities.

Modular Design for Scalability:

The machine's design should be modular, allowing manufacturers to scale operations based on demand. This flexibility would enable the addition of features or enhancements as technology advances or as production needs change.

The development of an automated folding and sewing machine for pleated front fabric is a crucial step toward addressing the challenges faced in manual sewing operations. By incorporating precision folding mechanisms, integrated line drawing systems, automated sewing units, real-time monitoring, user-friendly interfaces, and modular designs, manufacturers can significantly enhance efficiency, accuracy, and quality in garment production. This innovation not only promises to streamline the pleating process but also positions manufacturers to meet the evolving demands of the apparel industry, ultimately leading to improved competitiveness and sustainability.

### C. Automated Folding and Sewing Machine Design for the Pleat Feature on the Front

Following the study of the requirements, the design of an automated folding and sewing machine is required to feed the fabric sheet to the machine. After the automated machine feeds the fabric sheet, it requires holding the fabric sheet at high precision and then moving it to the single needle sewing machine. The speed of the sewing process from the single needle sewing machine should synchronize with the moving speed and sewing speed.

The mechanism of moving the fabric sheet relies on the gripper and its related grabbing mechanism. The speed of the gripper moving mechanism and sewing control mechanism are connected to the programmable logic controller (PLC). The PLC programs the moving speed and sewing speed. The sewing machine requires to use of a decoder to give a signal to the PLC for the sewing up and down cycle. Additionally, the PLC gives the signal and controls the sewing work of the sewing machine. PLC likes the main brain to control the

PLC system in Fig. 5 has the basic functional components of a processor unit, memory, power supply unit, input/output interface section, communications interface and programming device. The processor unit or central processing unit (CPU) is the unit containing the microprocessor and this interprets the input signals and carries out the control actions, according to the program stored in its memory, communicating the decisions as action signals to the outputs.

The power supply unit is needed to convert the main A.C. voltage to the low D.C. voltage (5 V) necessary for the processor and the circuits in the input and output interface modules.

The programming device is used to enter the required program into the memory of the processor. The program is developed in the device and then transferred to the memory unit of the PLC. The memory unit is where the program is stored that is to be used for the control actions to be exercised by the microprocessor, and data is stored from the input for processing and for the output for outputting.

The input and output sections are where the processor receives information from external devices and communicates information to external devices. The communications interface is used to receive and transmit data on communication networks from or to other remote PLCs. It is concerned with such actions as device verification, data acquisition, synchronization between user applications, and connection management.

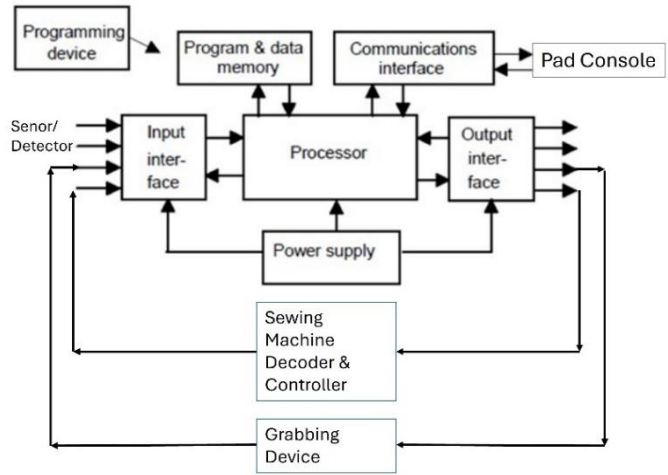

**Figure 5: Electricity Circuit and Control Diagram**

The design of the automated machine is required to grasp the fabric to pull the sewing machine for sewing the fabric after the folding process. The pulling force and its mechanism are required to pull the fabric to the Automated Folding and Sewing Machine for Pleated Pants Assembly in Fig. 6.

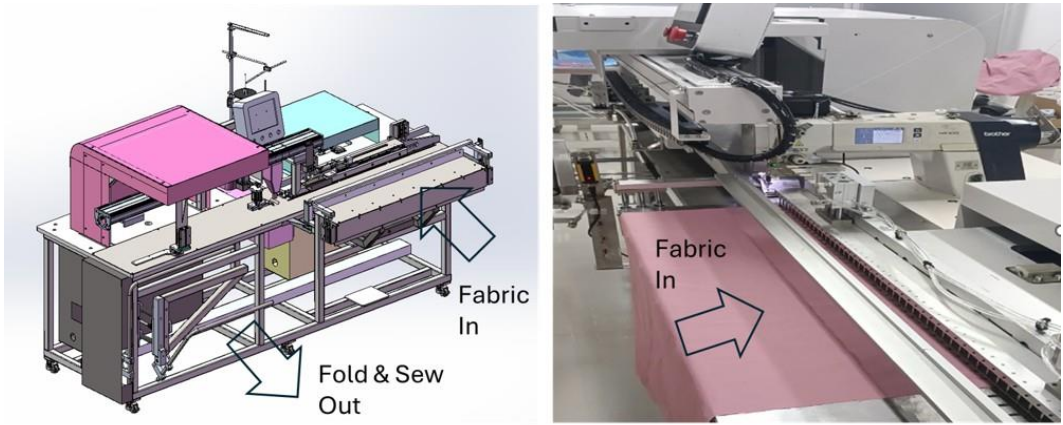

**Figure 6: Design of Automated Folding and Sewing Machine for Pleated Pants Assembly**

The automated machinery can perform the folding and sewing operation for pleated pants at high efficiency and high consistent quality.

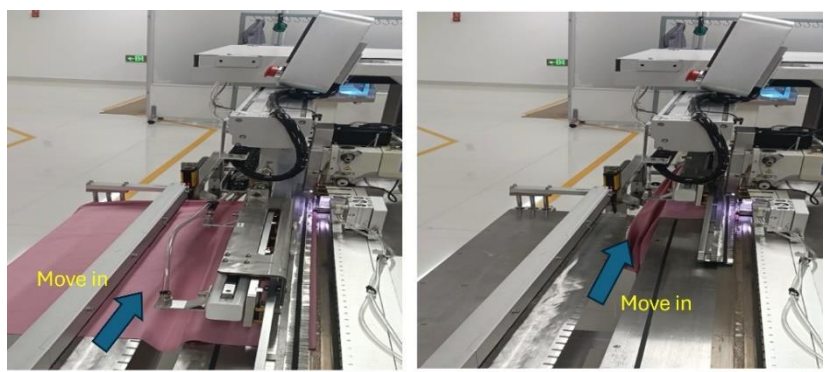

**Figure 7: Fabric piece moves in the sewing module for assembly**

In Fig. 7, the fabric piece can be moved in the sewing module by the pulling mechanism. The pulling device and mechanism can hold a fabric with high precision and parallelism without uneven flatness. The design of the moving mechanism for the fabric piece can satisfy the requirements for assembly.

The required driving force from the motor is required to overcome the friction between the fabric and the plate. The plate is a steel plate that keeps the parallelism with the moving gripper. The fabric is of slight weight. Its friction is near zero. The experiment has measured the resisting force from friction using by pull gauge. The pulling force of the fabric piece is driven by the pull gauge. The test result shows the nearly zero force of the fabric piece as shown in Fig. 8.

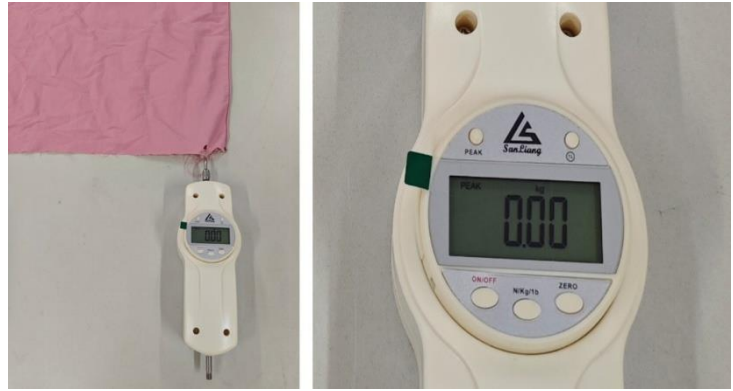

**Figure 8: Experiment of fabric piece fraction test**

The driving motor and mechanism can drive two forces output. One is the pressing force, which is a negative normal force in the vertical direction of the schematic diagram, $-F_N$, as shown in Fig. 1.

$$F_m = F_r - \mu F_N \qquad (2)$$

where $F_m$ is the moving force, $F_r$ is the driving force from the motor and $\mu F_N$ is the fraction.

Hence, the motor power is 750W, Torque of the motor is 2.5Nm. To calculate load torque, multiply the force (F) by the distance away from the rotational axis, which is the radius of the gear (r).

$$T = F * d \qquad (3)$$

where T = torque, F = force, d = distance

$T_m = F_r * d$ , $F_r = \frac{T_m}{d}$, $T_m = 2.4Nm$, $F_r = \frac{2.4Nm}{0.2m}$ which $d$ distance is 0.2m from the motor shaft)

$F_r = 12N$

Referring to the formula (2), the fraction, $F_N \cong 0$ , the moving force, $F_m = F_r - \mu F_N$ is shown below:

$F_m = F_r - \mu F_N$ ; $F_m = 12N - 0 = 12N$

The driving force 12N can drive the fabric sheet without a problem. The sewing speed is controlled by the PLC controller. The decoder and sensor can give a signal to the PLC controller. The design of an automated folding and sewing machine can successfully achieve the result.

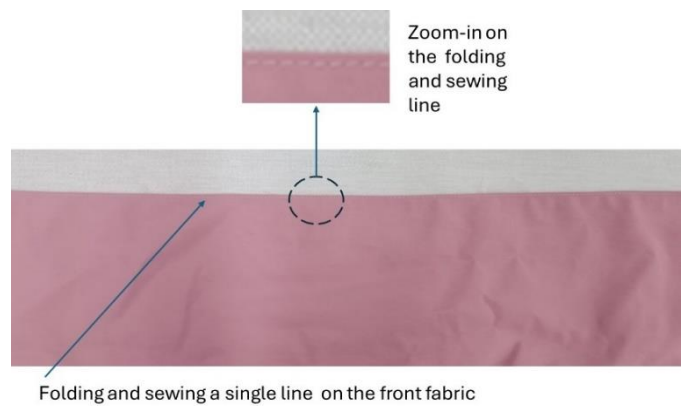

**Figure 9: Photo of folding sewing a single line on the front fabric**

The output of the sewing line within 1 mm of the folding edge or any specific requirement can achieve the appearance requirement, flatness and tolerance in Fig. 9.

## V.　CONCLUSION

The development of an automated folding and sewing machine for pleated pants represents a significant advancement in addressing the inefficiencies inherent in manual sewing processes. The challenges posed by labour-intensive, skill-dependent operations that often result in variability and inconsistency have been effectively tackled through this innovative solution.

The automated system incorporates several key features designed to enhance precision and efficiency. A precision folding mechanism ensures accurate alignment and consistent folding, while an integrated line drawing system eliminates the need for manual marking, thereby saving time and improving accuracy. The automated sewing unit, equipped with advanced technologies like programmable patterns and automatic tension control, guarantees precise stitching along the folded pleat, maintaining high-quality output.

Central to this system is the programmable logic controller (PLC), which acts as the command center, managing speeds, sewing patterns from the sewing machine, and synchronization between the moving and sewing processes. This ensures smooth operation and adaptability to different production needs. The design also includes real-time monitoring and feedback systems, allowing for immediate adjustments to maintain quality standards.

Pull force analysis was crucial in determining the optimal motor specifications, ensuring that the fabric is moved without distortion or misalignment. Experimental results confirmed that a motor with specific power and torque requirements could handle the fabric effectively, achieving the desired precision.

The successful testing of this system demonstrates its ability to meet stringent appearance requirements for flatness and tolerance, producing consistent and high-quality pleated pants. This innovation not only enhances efficiency and reduces production costs but also elevates product quality, positioning manufacturers at the forefront of the apparel industry.

In summary, this automated machine represents a leap forward in garment production, offering increased productivity, reduced variability, and improved product consistency. Its impact extends beyond immediate operational benefits, promising future enhancements and scalability to meet evolving industry demands.

**Benefits: labour operating time reduction**

The automated folding and sewing machine can eliminate the manual marking process in the whole process. The standard time and efficiency comparison table is shown below in Table 1.

### TABLE I: STANDARD LABOUR TIME COMPARISON

| Process | Involved Operator Standard Labour Time (seconds) | Involved Machinery Standard Time (seconds) | Standard time (second /piece) |
|---|---|---|---|
| *Manual Process* | | | |
| (a) Marking line on fabric sheet | 21 | 0 | |
| (b) Folding & Sewing | 96 | 96 | |
| | | | |
| *Manual Total:* | **117** | **96** | (a)+(b)= 117 |
| *Automation Process* | | | |
| (a) Loading the fabric sheet to the Machine | 6 | 0 | |
| (b) Folding & Sewing | 0 | 25 | |
| (c) Unload fabric sheet | 2 | 0 | |
| *Automation Total:* | **8** | **25** | (a)+(b)+(c)= 33 |
| **Comparison** | **Manual Process** | **Automation Machinery in Process** | **Improvement %** |
| **Std. Labour Time** | 117 | 8 | 93% <br> = (117sec- 8sec) / 117sec |
| **Std. Machinery Time** | 96 | 25 | 73% <br> = (96sec- 25sec) / 96sec |
| **Total Output Rate (sec/pc)** | 117 | 33 | 72% <br> = (117sec- 33sec) / 117sec |

The standard labour time is improved to 93%. The operator can assign 2 machines to get further enhancement. The standard machinery time is improved to 73% which shows the machinery can help it more. The total output rate (sec/pc) is shown the shortening the cycle time from 117sec/pc to 33sec/pc. The efficiency is enhanced by 72% due to the automated folding and sewing machine.

**Future Insight**

To analyze the design elements of front side and long pleated pants and understand their influence on fashion adoption, focusing on consumer preferences and market trends. The following methodology is for statistical analysis in apparel manufacturing.

**Data Collection:**

1. **Sources:** Gather data from fashion industry reports, sales figures, social media trends, and consumer surveys.
2. **Variables:** Consider factors such as fabric type, colour, number of pleats, hem style, target demographic (age, gender, income), and sustainability practices.
3. **Methods:** Conduct surveys or interviews with consumers who have purchased pleated pants to understand their preferences. Analyze sales data from retailers.

**Statistical Methods:**

1. **Descriptive Statistics:** Provide an overview of current trends in pleated pants designs and popularity.
2. **Inferential Statistics:** Use regression analysis to predict future trends based on historical data.
3. **Correlation Analysis:** Identify relationships between design elements (e.g., number of pleats) and sales or consumer satisfaction.
4. **Cluster Analysis:** Group consumers based on their preferences for specific pleated pants features.

**Considerations:**

1. **Bias Mitigation:** Ensure diverse sampling across regions and demographics to avoid bias.
2. **Time Frame:** Use recent data to account for rapidly changing fashion trends.
3. **Visualizations:** Utilize charts and graphs to identify patterns, such as scatter plots showing correlations between pleat number and sales.

**Limitations:**

- Statistical analysis may not fully capture subjective aspects of style and aesthetics crucial in fashion decisions.
- Combining quantitative data with qualitative insights (e.g., interviews) could provide a more comprehensive understanding.

This statistical analysis is planned to derive meaningful insights into the design elements influencing the adoption of pleated pants as a fashion style. By focusing on front side features and long pleats, the study seeks to inform design decisions and market strategies, ensuring alignment with consumer preferences and current trends.

## REFERENCES

[1] Kong, R. W. M., Liu, M., & Kong, T. H. T. (2024). Design and Experimental Study of Vacuum Suction Grabbing Technology to Grasp a Fabric Piece. OALib Journal , 11, 1-17. Article e12292. https://doi.org/10.4236/oalib.1112292

[2] Kong, R. W. M., Kong, T. H. T., & Huang, T. (2024). Lean Methodology For Garment Modernization. INTERNATIONAL JOURNAL OF ENGINEERING DEVELOPMENT AND RESEARCH, 12(4), 14-29. Article IJEDR2404002. http://doi.one/10.1729/Journal.41971

[3] Kong, R. W. M., Kong, T. H. T., Yi, M., & Zhang, Z. (2024). Design a New Pulling Gear for the Automated Pant Bottom Hem Sewing Machine. International Research Journal of Modernization in Engineering Technology and Science, 06(11), 3067-3077. https://doi.org/10.56726/IRJMETS64156

[4] Kong, R. W. M., Ning, D., & Kong, T. H. T. (2025). Mixed-Integer Linear Programming (MILP) for Garment Line Balancing. International Journal of Scientific Research and Modern Technology (IJSRMT), 4(2), 64-77. https://doi.org/10.5281/zenodo.14942910

[5] Kong, R. W. M., Ning, D., & Kong, T. H. T. (2025). Innovative Line Balancing for the Aluminium Melting Process. International Journal of Mechanical and Industrial Technology, 12(2), 73-84. https://doi.org/10.5281/zenodo.15050721

[6] Prof. Dr. Ray Wai Man Kong, Ding Ning, & Theodore Ho Tin Kong. (2025). Innovative Line Balancing for the Aluminium Melting Process. International Journal of Mechanical and Industrial Technology, 12(2), 73–84. https://doi.org/10.5281/zenodo.15050721

[7] Kong, R. W. M., Ning, D., & Kong, T. H. T. (2025). Line Balancing in the Modern Garment Industry. International Journal of Mechanical and Industrial Technology, 12(2), 60-72. https://doi.org/10.5281/zenodo.14800724

[8] Kong, R. W. M., Ning, D., & Kong, T. H. T. (2025). Innovative Vacuum Suction-grabbing Technology for Garment Automation. In K. M. Batoo (Ed.), Science and Technology: Developments and Applications (Vol. 6, pp. 148-170). BP International. https://doi.org/10.9734/bpi/stda/v6/4600

[9] Kong, R. W. M. (2025). INNOVATIVE AUTOMATED STRETCH ELASTIC WAISTBAND SEWING MACHINE FOR GARMENT MANUFACTURING. International Research Journal of Modernization in Engineering Technology and Science, 7(3), 7347-7359. https://doi.org/10.56726/IRJMETS70275

[10] Kong, R. W. M., Ning, D., & Kong, T. H. T. (2025). AI Intelligent learning for Manufacturing Automation. International Journal of Mechanical and Industrial Technology, 13(1), 1-9. https://doi.org/10.5281/zenodo.15159741

[11] Ray Wai Man Kong, Ding Ning, & Theodore Ho Tin Kong. (2025). AI Intelligent learning for Manufacturing Automation. International Journal of Mechanical and Industrial Technology, 13(1), 1–9. https://doi.org/10.5281/zenodo.15159741

[12] Kong, R. W. M. (2025). AI Magnetic Levitation (Maglev) Conveyor for Automated Assembly Production. International Journal of Mechanical and Industrial Technology, 13(1), 19-30. https://doi.org/10.5281/zenodo.15599657

[13] Kong, R. W. M. (2025). Woven Air Permeability Textile Fabric for Garment Automation. International Journal of Mechanical and Industrial Technology, 13(1), 31-45. https://doi.org/10.5281/zenodo.15806556

[14] Kong, R. W. M. (2025). Innovative Technology Strategy in Hong Kong and China of the Asia Industries. *International Journal of Social Science and Humanities Research*, *13*(3), 103-116. https://doi.org/10.5281/zenodo.15911676

[15] Domskiene, J., Mitkute, M., & Grigaliunas, V. (2023). Sewing and adhesive bonding technologies for smart clothing production. International Journal of Clothing Science and Technology, 35(4), 581–595. https://doi.org/10.1108/IJCST-02-2022-0028